# Reply to Comments on Neuroelectrodynamics: Where are the Real Conceptual Pitfalls?

Dorian Aur[1]

*Abstract: The fundamental, powerful process of computation in the brain has been widely misunderstood. The paper [1] associates the general failure to build intelligent thinking machines with current reductionist principles of temporal coding and advocates for a change in paradigm regarding the brain analogy.* Since fragments of information are stored in proteins which can shift between several structures to perform their function, the biological substrate is actively involved in physical computation. The intrinsic nonlinear dynamics of action potentials and synaptic activities maintain physical interactions within and between neurons in the brain. During these events the required information is exchanged between molecular structures (proteins) which store fragments of information and the generated electric flux which carries and integrates information in the brain. The entire process of physical interaction explains how the brain actively creates or experiences 'meaning'. This process of interaction during an action potential generation can be simply seen as the moment when the neuron solves a many-body problem. A neuroelectrodynamic theory shows that the neuron solves equations rather than exclusively computes functions. *With the main focus on temporal patterns, the spike timing dogma (STD) has neglected important forms of computation which do occur inside neurons. In addition, artificial neural models have missed the most important part s*ince the real *super-computing power of the brain has its origins in computations that occur within neurons.*

Intraneuronal information processing at the sub-cellular level is critically important to generate cognitive functions [2]-[8]. Cognitive abilities and behavioral changes are often explained solely based on the variability of temporal patterns though many scientists have well perceived the relationship between molecular signaling and 'neural code'. Interpreted as a natural variation of 'neural code' [9][10] the continuous change of temporal patterns hides the fundamental process of computation. A few researchers have understood that information processing at the sub-cellular level represents itself a 'neural code'. Any realistic model of computation has to show how information is transferred, processed, deleted or integrated in the brain. Since *'cognitive' computation cannot be reduced to information communication* between neurons, several models and critical comments in [11] fall short.

It is time to ask ourselves; where are the real conceptual pitfalls? Too often, the study of artificial neural networks (ANN) has proposed developments akin to studying the brain. Without understanding the role played by spikes in biological neurons, the research of ANNs has moved too fast to an analysis of neural ensemble. We have shown in [1] that artificial neurons which mimic temporal patterns (spiking neurons, weight type connection) are abstract mathematical models which fully compress information instead of processing information.

[1] Stanford University, Palo Alto, CA, USA
e-mail: DorianAur@gmail.com

In addition, the focus on solving equations is admirable. However, there is a foundational issue that must be addressed. Specifically, what do these equations describe rather than finding a 'rigorous' mathematical solution for some arbitrary equations. As Kolmogorov pointed out "it is not so much important to be rigorous as to be right". *The behavior of many natural systems including the brain cannot be well approximated using linear models or satisfy integrability conditions*.

**The Spike is Neither Digital nor a Symbol**

The simplest, realistic model of an action potential can be generated by the dynamics and interaction of electric charges (N-charge dynamics) [2][4]. The N-charge dynamics during an action potential is mathematically equivalent to a many-body problem [7]. In order to include meaningful changes of charge density experimentally observed [2][3][6][7], every action potential can be seen as a brief moment (1ms) when the neuron 'solves' at least a classical N-body problem. Either in a quantum approach of many-body physics [13] or in a classical framework of N-body dynamics [14] the difficulties of solving the problem of interacting charges are well known. Max Born well known for the development of quantum mechanics suggested that "Nature fortified herself against further advances behind the analytical difficulties of the many-body problem".

Simultaneous firing of action potentials (synchrony) in a neuronal ensemble increases the interaction of many- body systems and implicitly boosts the computational power [7]. The well accepted model of 'activation function' that describes artificial neurons simulated on Turing Machines does not approximate the process of 'solving equations'. Therefore, the popular claim that current artificial neurons and neural networks are 'realistic' models of their biological counterparts is misconstrued and untrue.

The neuroelectrodynamic model challenges the reductionist philosophy and shows that current models of computation, including the Turing model are a subset of a larger picture. Electrical events can be seen as moments that increase physical interaction. Since fragments of information are stored in proteins [21][7] which can shift between several structures to perform their function, then the biological substrate becomes an active part of the many-body physics and is directly involved in computation[2][7][12]. In [1] the many-body problem takes the Hamiltonian form. Therefore, equations (Eq-1 – Eq.5) suggest that recorded brain rhythms are *physically built up from a sub-cellular level through electrical interaction and regulated from a genetic level*. While other related aspects were previously presented [7] these five equations provide just a basic model of regulated interactions as has been already stated in [1]. With Hamiltonian equations written in [1] we follow theoretical ideas developed by great mathematicians and physicists. The integrability property is a simplification, like linearity or homogeneity assumptions.

Why do interpretations in neuroelectrodynamic (NED) theory generate a critical assessment of the temporal computing machine? The frequency of action potentials (APs) generated by neurons is correlated with different events; however, *it does not mean that such events are 'encoded' in the brain using the firing rate or any other 'temporal code'* (e.g. interspike interval).

**Experimental evidence: Why do neurons spike?**

The neuroelectrodynamic model was tested by experiment; however the reliance on false observations still remains an issue in neuroscience. The stereotyped action potential (AP) is just an appearance. Since *single-electrode recordings are inadequate to capture spatial propagation of a spike*, such fast modulation that occurs during AP generation (1ms) is completely ignored in current recordings [3][7]. Occasionally, a few electrophysiologists have observed changes in APs waveforms. They recognized these changes, unfortunately when they tried to understand what was really behind this phenomenon, they failed almost completely [20] (see explanations in [2]-[7]). Too many have ignored this phenomenon, the interpretation of small modulations of action potentials [3][5][6] is not 'carved' on the axonal branches.

The stereotyped action potential (AP) is just an appearance and is similar to the belief of a fixed, flat, Earth, a popular myth 500 years ago. Experimental data show patterns during action potential generation [2][3][7] which point to a sub-cellular level of information processing. The all-or-none action potentials are fast events; however they are not digital signals [28]. The meaningful change of spike directivity [6] contains the basic mystery of neuroelectrodynamics. This 'faithful conduction' [28] during APs represents a complex process of interaction when *information is electrically exchanged between molecular structures (proteins) which store fragments of information and the generated electric flux which carries information*. Every neuron 'speaks' in less than a millisecond during action potential generation. The action potentials are the meaningful 'words'.

*Electrical interactions that occur within neurons, inside the brain cannot be approximated by temporal patterns [3][7]. A comparative statistical analysis shows that electrical patterns approximated by spike directivity convey far more information regarding presented images than the firing rate or interspike intervals* [3]. This experimental result proves that temporal patterns don't provide a reliable approximation of electrical interactions. Importantly, the semantics are hidden in spatial modulation of action potentials which suggests that the *semantics are 'built' and 'integrated' into the cognitive level [3][6] through electrical interactions* [7][19] when *electrical events* are generated in *neuronal ensembles*.

Lord Adrian and many leading pioneers including Sir Alan Hodgkin could not envision that: (i) The transient electrical activity and temporal patterns cannot 'hold' fragments of information; (ii) Our memories need a more stable, non-volatile support at a molecular level (e.g. proteins); (iii) Simple cells have specialized into neurons, that are densely packed and generate *electric events at different scales to integrate information in the brain* [15]. Therefore, as a whole, *the entire brain is the 'computing' machine* and the NED framework extends Tononi's explanation regarding information integration [15] by including the relationship with biological substrate.

Unfortunately, too many took for granted Adrian's observations and hypothesized that neurons behave as metronomes. Lord Adrian can be easily excused since in the 1920s the brain research was in the early stages of paradigm development, there was no theoretical knowledge about computation or complex molecular signaling and no difference between information communication and computation.

Every model is able to characterize only a few properties of real physical phenomena. We didn't' learn how to fly by copying the morphology of birds' feathers. We had to understand the *physical principle* of winged flight. The inconvenient truth is that in the last sixty years we didn't learn how the mind emerges from temporal patterns since the entire framework *was attached to a false hypothesis (digital spike). Single-electrode recordings and fast propagation of action potentials were the origins of false observations*.

Inevitably, in science, any dogmatic view which fails to provide good explanations disappears sooner or later. The vagaries of observations of neural 'digital' events on a millisecond time scale and statistical method of thinking do not bring us any closer to the secrets of information processing in the brain. The convention of *the digital action potential is our mistake* which has artificially separated electrical events (e.g. APs) from molecular signaling [1] [17]- [19]. A simple observation of meaningful spatial modulation of APs [3][6] has a powerful far-reaching impact in neuroscience, neurology and computer science. Temporal patterns indicate solely when electrical events occur in neurons which partially characterize any electric event (e.g. *without disclosing "what information was processed" or "what information was electrically communicated"*). Therefore, the analysis of various time scales (firing rate, ISI, spike timing dependent plasticity) represent just a fraction of information needed to characterize electrical communication and not the entire process of computation.

Since memories are stored in proteins [21] (non-volatile structures) then a spatial modulation of spikes (see spike directivity) relates molecular computations with information transmitted during AP generation [19]. The similitude between APs and digital signals is irrelevant in terms of information processing. Both events may show similar shapes, however, *the intrinsic nature of computation is completely different*. Adding spikes or computing the firing rate to analyze behavioral semantics [6] or object recognition [3] *incorrectly simplifies* the 'neural code'. Therefore, theoretical constructs regarding Bayes theorem, nonlinear dynamics, have no real value in understanding how information is processed if they are attached to a false hypothesis (digital spike). In fact, many controversies, contradictions (e.g. temporal coding myths, concept cells, grandmother cells [7]) are generated by the 'most fundamental idea' of temporal coding [7]. Since *temporal patterns do not provide a reliable approximation of computations that occur within neurons [1][19],* then the temporal computing machine represents an obsolete model.

Borrowing terms from physics regarding 'open' brain are unfortunate since the research drifts away from the traditional concept. While neurons can be considered open systems since they share directly matter, energy and information with their surroundings [7] the extension of such physical concepts to the entire brain would be difficult to pursue. The receptors for the senses are in general not located inside the brain and during normal brain function neither matter nor energy is shared in a physical sense with the environment. In addition, thinking or reasoning, the cognitive (conscious) state can be maintained in the absence of external stimuli. *The process of thought is not simply driven by external sensory influences and sensory information may not necessarily be used in this process.* Therefore, the intact brain is not so physically 'open' to significantly leak energy, matter or information during thinking or reasoning.

The explanation is simple; it doesn't involve the 'open' brain and was put forward in neuroelectrodynamics. The brain (the 'thinking machine') has previously accumulated and stored information inside the biological substrate. Since our memories have a non-volatile support at a molecular level inside neurons within proteins, the generated electrical interactions (e.g. action potentials) dynamically 'read' and integrate the required information in the brain. If extracted from neuronal substrate, proteins cannot preserve meaningful information. The existence of nonlinear dynamics throughout the intra-cellular environment generates *an active biological substrate* [22] which provides a different explanation for the generation of action potentials [1]. Therefore, t*he thinking machine does not solely rely on sensorial information. This is a major qualitative change of paradigm introduced by the new explanatory approach from neuroelectrodynamic theory.* The model of regulated (electric) interactions determined by a specific biostructure (e.g protein) represents a more powerful framework for information integration than the one revealed by STD.

**The Systemic Model**

In addition, understanding the NED model requires the grasping of systemic approach. Many aspects related to metabolic processes, the effect of neurotransmitters and molecular computations (proteins synthesis, transcriptional-translational processes, enzymatic reactions) are included in the other two interacting loops represented in Fig 1, for details see also [7]. That's the level where dissipative structures can be correctly defined in terms of Prigogine's formal concept of self-organization. These changes in the organization that occur at a molecular (sub-cellular) level reshape charge densities within biological substrate and can be modeled as interaction terms in the Hamiltonian of the system.

The transient electric events (e.g. action potentials) integrate existent information from molecular structures (e.g. proteins) on a system level. Importantly, the entire neurochemistry [24] and metabolic pathways are needed to maintain and regulate the fundamental process of electrical interaction in the brain. The systemic approach in neuroelectrodynamics makes the 'whole' from the interaction of the parts. *This is a more substantial paradigmatic shift and the neuroelectrodynamic model becomes an*

*integral part of systems biology [16].* Having everything in a single loop (e.g. temporal coding) would indeed follow a reductionist philosophy.

In addition, the *process of computation in the brain emerges from physical interactions,* not from resulting dissipation when information can be deleted (see the Landauer's principle [7][23]). Given physical implementation, an energy cost is unavoidable; however, the process of dissipation which occurs in many other systems *is not a condition to generate semantics or cognitive abilities*. Since fragments of information are stored within molecular structure (proteins) in neurons a minimalist model of 'access to memory' to 'read' information through physical interaction is required. The Hamiltonian formalism can well model nonlinear interactions (e.g. the N-body problem).

The fundamental principle of computation in the brain is based on specific interactions. The elegant simplicity of controlling physical interaction in the brain and neurons is meaning-making [3][6]. In [1] we tried to find a simple, yet a non-trivial model to describe the non-mysterious nature of such interactions. The Hamiltonian structure in [1] extended to infinite dimension case models the interaction of electric field with charges, resulting resonant regimes or chaotic dynamics. These interactions intrinsically exhibit many features such as parallelism, fuzziness, fractional dynamics and emergent phenomena [26]. In addition, for more than two decades the development of reaction diffusion models (dissipative systems) and Hamiltonian systems was separated. Elgart and Kamenev have recently established the correspondence between reaction-diffusion systems and Hamiltonians [27]. *A reaction-diffusion system (dissipative system) can be described by the Hamiltonian action and the resulting system can be modeled and analyzed based on corresponding (non-integrable) Hamiltonians as in [1]*. The many-body problem as well the Hamiltonian formalism is not limited to a confined scale. *From a quantum (molecular dynamics) to classical systems (celestial mechanics), Hamiltonian models can approximate nonlinear interactions*. Feynman has presented the first convincing Hamiltonian framework for physical (quantum) computation a few decades ago [25]. The approximation in the action-angle form (Eq.2) generates extreme examples of the kind of behavior the brain can exhibit. There's no need to solve " 6M1010 equations" to understand that complex brain rhythms, resonant regimes (Eq.3) or generated chaotic dynamics (Eq. 4) represent this kind of 'solution'. The AP itself is a complex process of interaction where resonant regimes are present and can be evidenced within generated electrical patterns [2][3][7]. *In addition, what we can analytically solve, compute or simulate on Turing machines has no relevance and cannot be used as an argument to refute a natural model of computation*.

The NED model provides a methodical explanation on how the brain creates or experiences 'meaning'. The results indicate that 'cognitive maps' have their origins in interactions that occur inside many neurons which 'fire together' to electrically integrate information in the brain (see Fig. 5 in [3]). All these experimental results confirm previous theoretical work [31][8] and explain why a digital approximation of APs completely changes the nature of computation that leads to an incomplete model [32]. Without associated semantics, the entire algorithmic construct of temporal computing machine is meaningless.

While 'weak' interactions between neurons may be algorithmically approximated, 'strong' interactions that occur inside neurons during AP generation remained unmodeled. In a Turing model, macromolecular assemblies (proteins) which store fragments of information can represent the 'tape' and the electric (ionic) flow can be the 'head' which dynamically 'reads' or 'writes' information inside neurons [7][19]. However, the relevance of Turing model is questioned even in case of present-day computing [33] [34]. Indeed, any computing machine that follows a Turing model would be highly inefficient to simulate the activity of biological neurons and experience an increased slowdown. Since the *super-Turing computing power of the brain has its origins in these 'strong' interactions that occur inside neurons, current models have missed the most important part. Simply, Nature doesn't care if the N-body problem has analytical solutions [36] or can be simulated in real time on a Turing machine [37].*

In addition, we are not concerned about 'small imperfections' (e.g. dissipation) virtually, included references and any undergraduate textbook explain introductory topics on Hamiltonian mechanics (e.g. integrability). *In the last sixty years the studies of Hamiltonian systems have moved further to include the physics interesting points, which are beyond complete integrability*. With equations written in [1] we follow theoretical ideas developed by great mathematicians and physicists. For details see references in [1].

While previous models have attempted to represent Hamiltonians using Turing machines [35] the paper [1] shows that *the Hamiltonian model of interaction can represent itself a far more powerful model of computation*. Turing made an important step forward; however, there is no need to limit natural models of computation to Turing models. In this sense, the new framework of computation using interaction is universal in nature and provides a more general description of computation than the formal Turing model. In other words *God was unaware of Turing's work and has put forward a better model for physical computation in the brain*.

Once this chronic failure generated by temporal computing machine is accepted, further progress is possible. We have to admit that up today consciousness didn't emerge from STD or any other algorithmic model. *The entire solution regarding consciousness lies in 'strong' interactions developed within neurons.* The process of *computation by physical interaction [19] can compute all possible sets and functions [29] and generate itself a more powerful model of computation with a super-Turing behavior [30]*. Since fragments of information are stored inside neurons, then the regulated electrical interactions can bring information together; give rise to memory based experience which provides the basis of cognition and consciousness [7].

Indeed, the theory of mind and consciousness can have a computational framework [38] however, it is a non-algorithmic one. Even the 'non-computational' theory of consciousness presented by Sir Roger Penrose [39] (incomputable, in the Turing sense) becomes 'computational' within the new description. Once we have the right theory of computation [7] the unobservable mind [40] turns out to be observable from the smallest level. Remarkable, similar views regarding physical processes are lately shared by Terrence Deacon [41] and confirm our approach to understand the brain.

The neuroelectrodynamic model provides a new computational theory of mind, brings back the strength of physical laws which show how information is transferred, processed deleted or integrated in the brain [1][7]. There is no need to develop temporal coding frameworks to understand how the *mind dynamically emerges from the interaction of matter*. Physical models can explain better these phenomena. The new framework NED, provides clear explanations for many unsolved brain mysteries (e.g. mirror neurons, phantom limb, sparse coding), generated neurological disorders (e.g. seizure generation in epilepsy [42]) or even for the 'natural' variation of hypothesized neural code [9]. The theoretic model had to be changed to include experimental observations. At least three regulatory loops (see Fig.1) characterize biological substrate which generates brain rhythms underlying cognitive processes and consciousness

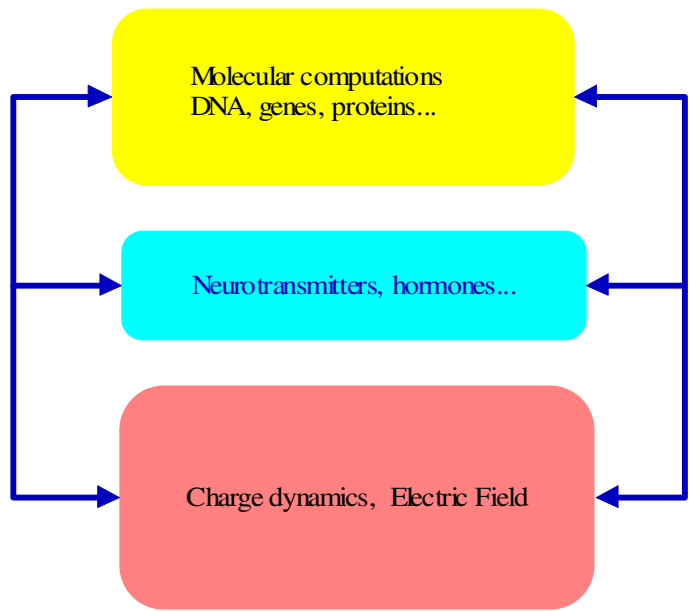

Figure 1: A schematic presentation of three regulatory loops that characterize biological substrate (adapted from [1])

**What will be Next?**

Without a doubt, as presented today 'neural' computation describes a model of communication between neurons rather the required physical model of computation. A common mix-up is to claim that 'spikes convey information', then hypothesize, they are digital events and limit the entire model to a digital communication of temporal patterns. The simple process of communication (either modulated or not) does not describe the entire process of computation. At least information storage, information integration has to be included in the model and related to changes that occur in biological substrate (e.g. proteins), see details in [1][7]. Otherwise, all temporal patterns disappear into 'thin air'.

Indeed, "physical realization" needs to be considered and *the simplicity underlying the physics of interaction is highlighted in neuroelectrodynamics* [7]. We are pleased to see that other scientists start to understand that information is not stored into 'thin air' (e.g. temporal patterns) and that fragments of information can be physically deleted in the brain. In fact this is the first compelling proof of understanding neuroelectrodynamics' principles. Does spike timing dogma or 'classical' artificial neural networks explain these processes which occur in the real brain? No, they have remained too 'artificial'. In the last sixty years, only a few models have included the relationship with biological substrate (see references in[1][7]).

Following the language of Thomas Kuhn, a new paradigm was introduced in [1] and related to requirements to build reliable thinking machines in [7]. The neuroelectrodynamic theory proposes a more pragmatic model for cognitive computation which extends Goldin- Wegner's paradigm of *interactive* computation and Tononi's model of integrated information. Within a systemic approach of many-body physics the NED theory includes the biological substrate of neurons (electric charges ,proteomics, genomics) as the fundamental basis of computation in the brain. Furthermore, STD can be seen as a particular case of neuroelectrodynamics, an approximation when APs are digital events and neurons behave as metronomes. One could try to record, perform statistics or mimic the occurrence of temporal patterns for a lifetime without learning anything about what they really represent.

The spike timing dogma has artificially amplified the gap between cognitive processes and physical realm. Given the focus on the boundaries between biological grounds and cognitive computation, general topics related to Hamiltonian dynamics (e.g. integrability) remain the Quixotian 'windmills' believed to be the malicious giants.

Only a few researchers have understood that *cognitive processes and emerging consciousness are grounded in continuous, regulated dynamics of physical interactions in the brain*. In addition, well known limitations are attached to the Turing model and computational theory needs to incorporate other forms of computation [1][19][33]. Even the basic question should be rephrased in [43], "Is the Turing framework the best model for machine intelligence?" This aspect and the required paradigm shift will be obvious in a few decades *since the fundamental, powerful process of computation in the brain has been widely misunderstood*.

**Acknowledgments**

Thanks to Nathaniel Bobbitt for excellent suggestions to improve the manuscript
.